\pdfoutput=1

\documentclass[10pt,twocolumn,letterpaper]{article}

\usepackage{cvpr}              

\usepackage{graphicx}
\usepackage{amsmath}
\usepackage{amssymb}
\usepackage{booktabs}
\usepackage{hyperref}
\usepackage{cleveref}

\usepackage{amssymb}
\usepackage{colortbl}
\usepackage{graphicx}
\usepackage{multirow}
\usepackage{color}
\usepackage{bbding}
\usepackage[linesnumbered,ruled,vlined]{algorithm2e}
\usepackage[super]{nth}

\definecolor{mygray}{gray}{.9}
\definecolor{light-gray}{gray}{0.72}

\definecolor{citecolor}{RGB}{34,139,34}

\makeatletter\renewcommand\paragraph{\@startsection{paragraph}{4}{\z@}
	{.35em \@plus1ex \@minus.2ex}{-.5em}{\normalfont\normalsize\bfseries}}\makeatother

\def\cvprPaperID{} 
\def\confName{CVPR}
\def\confYear{2023}

\begin{document}

\title{Unleashing the Power of Depth and Pose Estimation Neural Networks by Designing Compatible Endoscopic Images}

\author{
Junyang Wu\textsuperscript{1} ~ 
Yun Gu\textsuperscript{1} ~ 
\\
\textsuperscript{1}Shanghai Jiao Tong University \qquad \qquad 
}
\maketitle

\begin{abstract}


Deep learning models have witnessed depth and pose estimation framework on unannotated datasets as a effective pathway to succeed in endoscopic navigation. 
Most current techniques are dedicated to developing more advanced neural networks to improve the accuracy. 
However, existing methods ignore the special properties of endoscopic images, resulting in an inability to fully unleash the power of neural networks.
In this study, we conduct a detail analysis of the properties of endoscopic images and improve the compatibility of images and neural networks, to unleash the power of current neural networks.
First, we introcude the Mask Image Modelling (MIM) module, 
which inputs partial image information instead of complete image information, allowing the network to recover global information from partial pixel information. This enhances the network’s ability to perceive global information and alleviates the phenomenon of local overfitting in convolutional neural networks due to local artifacts. Second, we propose a lightweight neural network to enhance the endoscopic images, to explicitly improve the compatibility between images and neural networks.
Extensive experiments are conducted on the three public datasets and one inhouse dataset, and the proposed modules improve baselines by a large margin.
Furthermore, the enhanced images we proposed, which have higher network compatibility, can serve as an effective data augmentation method and they are able to extract more stable feature points in traditional feature point matching tasks and achieve outstanding performance.

\end{abstract}


\section{Introduction}

With development of precision medicine, minimally invasive surgery has become a major development direction in medicine. Endoluminal intervention aims to reach a lesion through the body’s cavity or lumen for biopsy or treatment. As a non-invasive intervention tool, endoscopes play an important role in endoluminal treatment. However, the pathways of lumen structure can be complex. How to design a suitable navigation algorithm to guide doctors to manipulate endoscopes to reach the target area is a clinical challenge. 

For visual navigation, depth and pose estimation is an important challenge. The depth information of endoscopic images can measure the distance between the endoscope and the organ wall to avoid collision damage between the instrument and the patient’s organs; pose information can perceive the position of the endoscope in real time and guide doctors to reach the target area according to the preoperative planned path. Moreover, robust pose and depth estimation techniques enable advanced applications like augmented reality and automated medical interventions, revolutionizing the field of endoscopy and pushing the boundaries of medical innovation.

Traditional multiview stereo method is a fascinating area of computer vision research that aims to reconstruct the three-dimensional structure of a scene from multiple images, e.g., structure from motion(SfM)\cite{sfm} and simultaneous localization and mapping(SLAM)\cite{chen2018slam}.
It leverage the power of multiple viewpoints to infer the depths and positions of objects in the scene. By analyzing the correspondences between points in different views and incorporating geometric constraints, traditional multiview stereo methods can estimate accurate depth maps and camera poses. 
However, due to the sparse and uneven distribution of feature points in endoscopy, traditional methods cannot achieve satisfactory results.

In recent years, with the development of deep learning, learning-based methods to estimate pose and depth have explored rapidly. 
These methods\cite{zhou2017unsupervised, bian2021unsupervised, monodepth2, watson2021temporal, ozyoruk2021endoslam, AF} fully take advantage of the fundamental principles of SfM and train the depth and pose neural network by minimizing the appearance difference between consecutive target and frames.

Although many works focus on designing more advanced networks for depth and pose estimation tasks, in endoscopic scenarios, we believe that in addition to considering more sophisticated networks, it is crucial to address the question of whether the original endoscopic images are compatible with current convolution neural network. 
Due to the presence of artifacts and sparse feature inherent to endoscopic scenes, the original endoscopic images may introduce much noise to the neural network. 
In addition, within the self-supervised framework, images serve not only as inputs but also as a source of supervisory signals for network training. Therefore, designing images that are better compatible for the network may potentially lead to improved performance across various neural networks.





In this paper, we focus on analyzing the limitations of endoscopic image characteristics in depth and pose estimation and explore the endoscopic images that are better compatible to convolution neural networks.
Compatible images to the neural network can unleash the power of CNN without significantly increasing the burden.

Concretely, we first introduce the concept of Mask Image Modeling (MIM) into endoscopic image deep pose estimation, enhancing the neural network's perception of global information. We observe that due to the local receptive field of convolutional neural networks, the depth estimation network may perceive specular artifact features as depth features, leading to local depth overfitting. To alleviate this phenomenon, instead of inputting the complete image, we propose partially masking pixel blocks to enable the network to estimate the depth of the invisible regions through contextual features, thereby improving the network's perception of global information. In addition, we discuss the optimal masking strategy for addressing the characteristics of artifacts in endoscopic images. In light of the artifact features in endoscopic images, we explore three masking strategies: random masking, uncertainty-guided masking, and artifact suppression masking, and conduct a comprehensive discussion on their effectiveness.

Second, owing to the stable characteristic of reconstructed image produced by MIM, we strike a balance between detail-rich original image and stable reconstructed image, in order to generate images that encompass both the richness of high-frequency features and the stability of low-frequency features.
Specifically, we design a lightweight enhanced network and feed both original image and reconstructed image, 
then constrain the generated images using downstream objective function, in order to genreate enhanced images that are well-suited for depth and pose estimation networks.


Our main contributions of this work can be summarized as follows:

\begin{itemize}
    \item Instead of developing more advanced networks, we conducted a detailed analysis of the properties of images. We analyzed the shortcomings of endoscopic images in depth and pose estimation tasks and improved the compatibility of images and neural networks by optimizing the input images.
    \item The concept of Mask Image Modeling has been introduced into the endoscopic depth and pose estimation algorithm, overcoming the challenge of local overfitting of neural networks for depth estimation. Furthermore, different mask strategies can overcome various artifacts in endoscopic sences.
    \item A novel enhancement network is proposed. The network is capable of adaptively selecting image detail features that are beneficial for downstream tasks and generating images with more stable features. The generated images have been verified to have advantages in multiple aspects.
\end{itemize}

\section{Related Works}

In this section, we review the related works on supervised depth and pose estimation and self-supervised depth and pose estimation.

\subsection{Supervised depth estimation}

Supervised depth estimation is a deep learning task that aims to predict the distance of each pixel in an image relative to the camera. It requires a large amount of labeled data to train a neural network that can directly output the depth prediction. 
The first supervised deep learning framework proposed for depth estimation \cite{eigen2014depth} introduces convolution network to estimate the depth and design scale-invariant loss.
Later works improve performance and efficiency of supervised depth estimation. 
\cite{laina2016deeper} consists of a series of residual blocks that learn to refine the depth map progressively from coarse to fine resolution. DORN\cite{DORN} discretizes the continuous depth values into a number of intervals and trains a deep convolutional neural network to perform ordinal regression, which predicts the order of the depth intervals.
DPT\cite{DPT} uses vision transformers as backbones and proposes a novel self-attention mechanism that incorporates positional encoding and relative position bias to capture long-range dependencies and local context.
These methods achieved excellent performance on datasets such as KITTI and NYU. 
However, due to the lack of large-scale annotated datasets for endoscopic scenes, it is difficult to apply fully supervised depth estimation to endoscopic scenarios. In this work, we focus on self-supervised depth estimation methods that do not require annotated data.

\subsection{Self-supervised depth and pose estimation}

To overcome the challenge of insufficient annotated data, researchers begin to explore the self-supervised depth and pose estimation using neural networks. 

As a pioneering work, \cite{zhou2017unsupervised} desgined two seperate network to estimate the depth and pose, respectively, and optimizing neural networks based on structure from motion framework. 
After that, \cite{bian2021unsupervised} proposes a geometry consistency loss to enforce scale-consistency across different frames and views. \cite{monodepth2} introduces full-resolution multi-scale sampling, which reduces visual artifacts and improves fine details by sampling pixels at their original resolution. \cite{watson2021temporal} uses a deep end-to-end cost volume based approach that adapts to the available sequence information.

After witnessing the success of depth and pose estimation in deep learning, researchers begin to explore the self-supervised depth in endoscopic scenarios.
Using deep-learning based dense SLAM, \cite{ma2019real}  achieved real-time dense reconstruction of colon images.
Endo-SfMLEarner\cite{ozyoruk2021endoslam} introduce spatial attention mechanism to improve the pose net. AF-SfMLearner\cite{AF} propose Appearance net to calibrate the brightness condition.

Rather than designing advanced network, in this work, we focus on analyzing the limitation of endoscopic image and contemplate what types of images can better unleash the power of neural network.


\section{Methodology} 

In this section, 
we first briefly summarize the preliminaries of the self-supervised monocular depth and pose estimation.
Then, based on the phenomenon of local overfitting in endoscopic images, we propose the Mask Image Modelling(MIM) module, which improves the generalization ability of depth network. 
Finally, we propsoe an enhanced network to strike a balance between low-frequency stable feature and high-frequency rich-detail information, in order to generate images with more rich and stable features.

\subsection{Preliminaries}

The goal of depth and pose self-supervised learning is to train depth and pose CNNs from unlabelled endoscopic videos. The main idea behind it is to use photometric consistency loss as the supervision signal.
Given consecutive frames $I_{t}$ and $I_{s}$ in a video, the depth of $I_{t}$ and the relative 6-DoF camera pose $P^{t-s}$ are estimated by the depth and pose CNNs, respectively. 
With estimaged depth and pose, we can synthesize the target image $I_{t}$ using the source image $I_{s}$ by differentiable warping as 

\begin{equation}
I_{t}^{'} = [K|0]P^{t-s}\begin{bmatrix} DK^{-1}I_{s} \\1  \end{bmatrix}
\end{equation}

Then the depth and pose CNNs are supervised by the photometric loss between the real target image A and the synthesized image B.

In practice, the photometric is usually the combination of L1 Loss and SSIM loss:
\begin{equation}
    L_p = \alpha \  SSIM(I_{t}, I_{t}^{'}) + (1-\alpha) |I_{t} - I_{t}^{'}|
\end{equation}

In addition, existing work incorporates a smoothness prior to regularize the depth map using the edge-aware smoothness loss:
\begin{equation}
    L_{s} = \sum_{p}(\exp^{-\nabla I_{t}(p)} \nabla D_{t}(p))^{2}
\end{equation}

Overall, the final objection function is :
\begin{equation}
    L = \beta L_{p} + \gamma L_{s}
\end{equation}
where  $\beta$, $\gamma$ are the loss weighting terms.

\subsection{Property of endoscopic images}

\begin{figure}[t!]
\centering
\includegraphics[width=\linewidth]{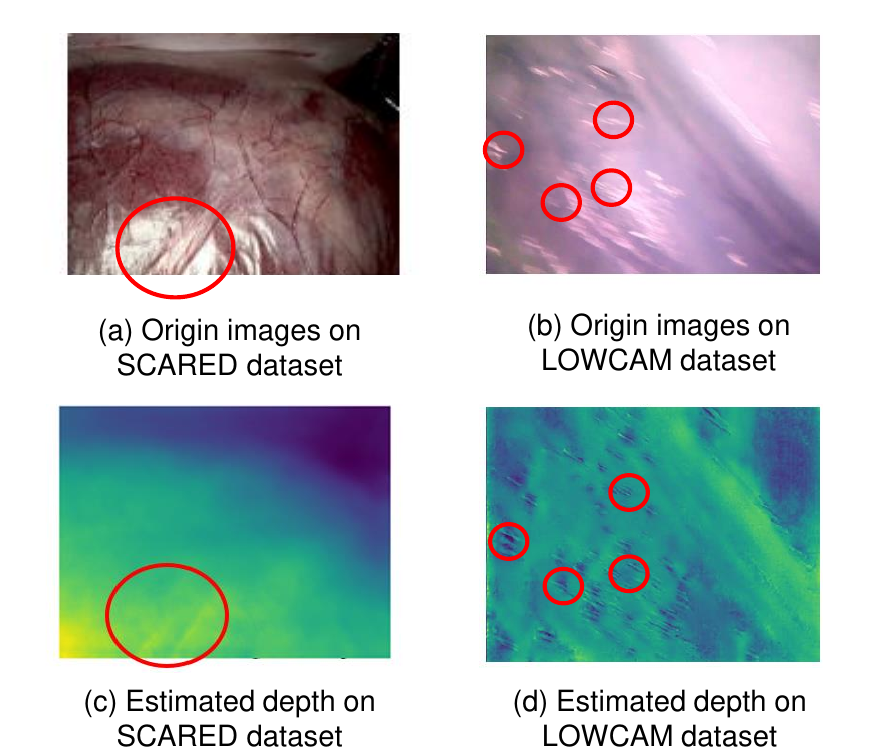}
\vspace{-1.5em}
\caption{The examples of locam overfitting.On the SCARED dataset, the depth network identifies the specular highlight area as a depth change area. On the LOWCAM dataset, the depth network mistakenly identifies specular features as depth features and ignores depth changes.}
\vspace{-.5em}
\label{fig:local_overfit} 
\end{figure}

One of the main characteristics of endoscopic images is that they have artifacts such as lighting, reflection, motion blur, bleeding, and noise, which means that not all the detailed features of endoscopic images are conducive to depth and pose estimation. Due to the local receptive field of convolutional neural networks, it may fit the features of artifacts as depth features. For example, as shown in \cref{fig:local_overfit}, the neural network considers the features of specular artifacts as depth features, but ignore the depth changes in the image itself.
Therefore, we hope that the network can have a larger receptive field to consider the global information of the image rather than being trapped in local pseudo-features.

\begin{figure}[t!]
\centering
\includegraphics[width=\linewidth]{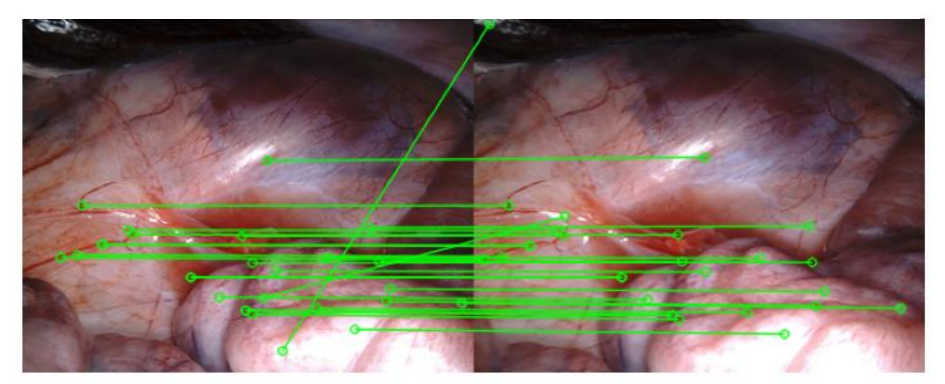}
\caption{The examples of keypoint mismatching. Adjacent endoscopic images are processed using SIFT to extract feature points and KNN algorithm for feature matching. A total of 22 pairs of matching points were extracted, with obvious mismatches present.}
\vspace{-.5em}
\label{fig:mismatching} 
\end{figure}

In addition, due to the sparse and uneven distribution of texture features in endoscopic images, traditional feature point matching-based pose estimation methods fail. We use the SIFT algorithm to extract feature points from two endoscopic images and use the KNN algorithm for feature matching. As shown in \cref{fig:mismatching}, we obtain 22 pairs of feature points, and obvious mismatches are produced.

Although feature points are not explicitly extracted based on deep learning methods, convolutional neural networks still extract low-level information based on image features. The characteristics of uneven distribution and sparsity in endoscopic images also cause chanllenges on deep learning methods.

Both of the above characteristics cause challenge of depth pose estimation in endoscopic images. To overcome them, this paper proposes two innovative improvements to make endoscopic images more suitable for depth pose estimation tasks. 
Firstly, our proposed Mask Image Modelling(MIM) strategy enhances the network’s ability of perceiving global information and forbids local overfitting.
In addition, our proposed lightweight enhanced network generates images with more rich and stable features, enhance the correspondence between adjacent images.


\subsection{Mask image modeling in endoscopic images}

The local receptive field of convolution neural network may cause local overfitting and overfit the features of artifacts as deep features.
In order to increase the receptive field and enhance the network’s ability to perceive global information, we introduce the concept of MIM into depth and pose estimation.

\begin{figure}[t!]
\centering
\includegraphics[width=\linewidth]{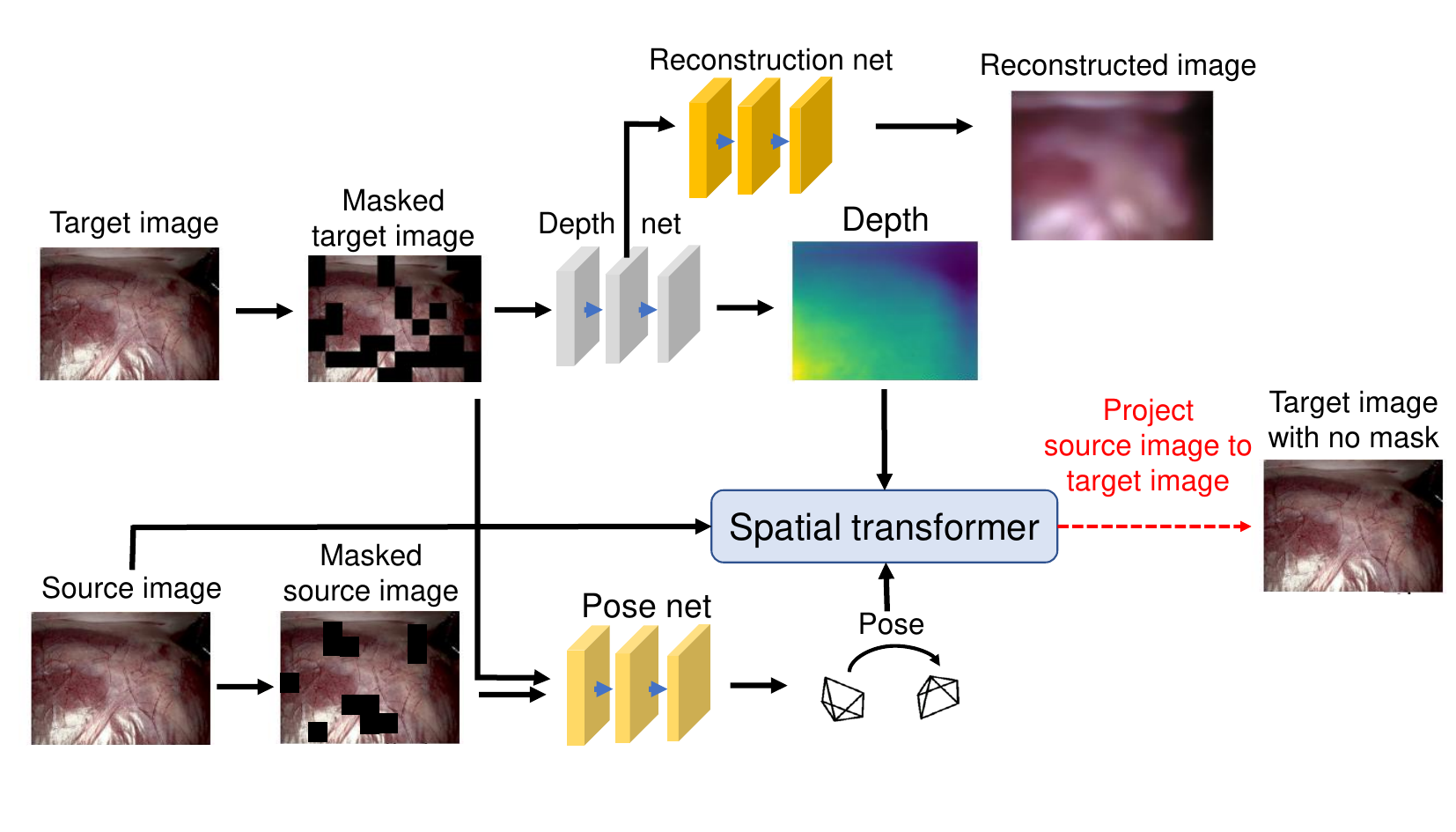}
\caption{The pipeline of MIM module. The input to the neural network is masked image, and the output and supervision signal is complete.}
\vspace{-.5em}
\label{fig:MIM} 
\end{figure}

As shown \cref{fig:MIM}, we design an asymmetric pipeline of self-supervised depth and pose estimation. 
Intuitively, the aim of MIM is to train the neural network to restore the complete depth map and complete image with the partial input information.
In order to complete the depth and pixel information of the invisible regions, the network needs to perceive contextual information, thereby obtaining more global information. 

Specifically, 
for forward, we first apply some mask strategies to the input image. Then, we feed the masked image to depth encoder to extract features. After that, our mask module has two branches: reconstruction decoder and depth decoder. For reconstruction decoder, the output is the estimated input image without mask. The depth decoder aims to estimate the depth of the input without mask. 
For backward, we keep the source image and target image without masking. Since the supervision signal is dependent on the source image and target image, our design aims to make our network estimate the whole image information with partial information, in order to improve the global perception ability of our network.
In addition to estimating depth, the loss function of the restoration network is the L2 loss between the output image and the origin image. This loss more directly constrains the network to recover complete global information from partial pixel information.

\IncMargin{1em}
\begin{algorithm} \SetKwData{Left}{left}\SetKwData{This}{this}\SetKwData{Up}{up} \SetKwFunction{Union}{Union}\SetKwFunction{FindCompress}{FindCompress} \SetKwInOut{Input}{input}\SetKwInOut{Output}{output}
	
	\Input{target image $I_{t}$, source image $I_{s}$, mask strategy $M()$, Depth net D(), pose net P(), reconstruction net R(), warp function W()} 
	\Output{estimated depth d, estimated pose p, reconstructed image $I_{t}^{'}$}
	 \BlankLine 
	 
	 \While{training}{ 
            $I_{t}^{M}$ = M($I_{t}$), $I_{s}^{M}$ = M($I_{s}$)\;
            
            d = D($I_{t}^{M}$), p = P($I_{t}^{M}$, $I_{s}^{M}$), $I_{t}^{'}$ = R($I_{t}^{M}$)\;

            warped image $I_{t}^{w}$ = W($I_{s}$, d, p)\;

            photometric loss $L_{p} = |I_{t}^{w} - I_{t}|$\;

            reconstruction loss $L_{r} = |I_{t}^{'} - I_{t}|$\;

            loss = $L_{p} + L_{r}$\;

            backpropagation\;
  
 	 } 
 	 	  \caption{Mask Image Modelling Pipeline}
 	 	  \label{alg:mim} 
 	 \end{algorithm}
 \DecMargin{1em} 

For mask strategies, we design different and analyze them respectively. 

\paragraph{Random mask} First, we simply choose random mask as our mask strategy. For the input image with size H$\times$W$\times$3, we first split it into several image patches, which size is p$\times$p$\times$3. The total number of patches is $\frac{H}{p} \times \frac{W}{p}$. 

\paragraph{brightness mask} 
Due to the adverse impact of the specular artifact on pose estimation, we suppress the speckle area of the endoscopic mask image reconstruction based on this perspective. Since the idea of mask image modeling can complete the mask area through contextual information, we hopes to block the specular area and restore the depth through the network’s global perception ability, thereby alleviating the negative effects brought by the specular artifact. 
Specifically, we follow \cite{} to detect and mask the specular highlight.

\paragraph{Uncertainty guided mask} 
Uncertainty is a important concept in computer vision. It is a measure of how confident a model is about its predictions or outputs. It can help us understand the limitations and reliability of the model, and make better decisions based on the model’s outputs.
We design a uncertainty-guided mask to improve the efficiency of mask training. Instead of masking randomly, we focus on areas that the neural network cannot estimate the depth confidently. 
Specifically, we first randomly mask input images, each masked images can produce a corresponding depth estimation, then we calculate the standard deviation as the uncertainty. Finally, we mask the areas with the highest uncertainty. 

In our experiment, we find that using brightness mask and uncertainty mask empirically works the best, and the detail experiment will be shown in Sec. XX

\subsection{Refined network}

\subsubsection{Properity of reconstruction images} 

\begin{figure}[t!]
\centering
\includegraphics[width=\linewidth]{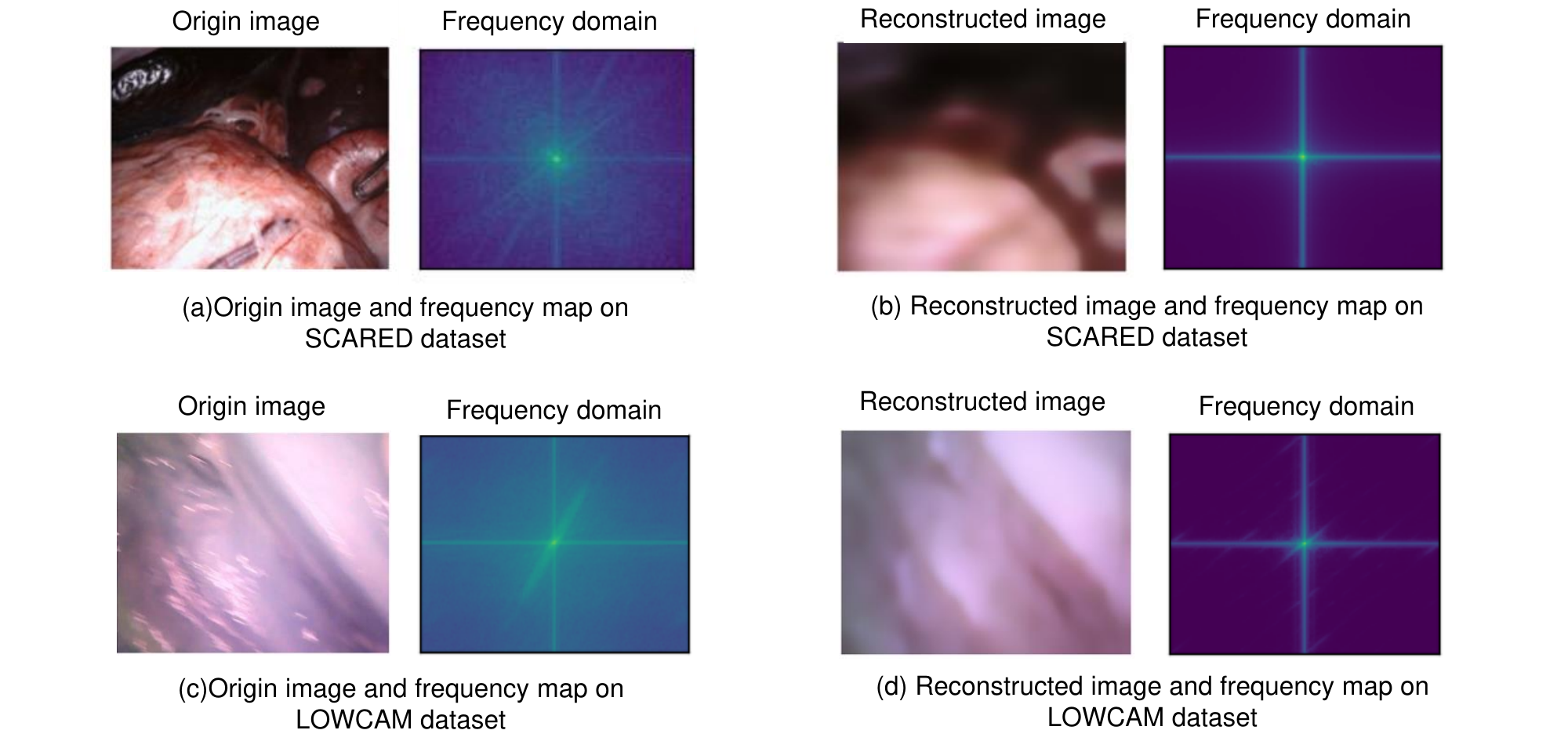}
\caption{Comparison of the properties of the reconstructed image and the original image. The reconstructed image has been smoothed based on global statistical features, resulting in a more low-frequency and stable overall signal, while the original image has richer details.}
\vspace{-.5em}
\label{fig:origin_image_vs_recon} 
\end{figure}

In MIM module, besides estimating the depth, the reconstruction decoder reconstruct the whole image. As shown in \cref{fig:origin_image_vs_recon}, the reconstructed images have some properties: 

Compared to the original image, the reconstructed image discards detailed information but retains the overall structural information. In endoscopic images, we hope to retain stable structural features and stable detailed features (such as edges and blood vessels), while discarding unstable detailed features (such as light spot artifacts). Therefore, we hope to strike a balance between the original image and the reconstructed image.

\subsubsection{refined network}

\begin{figure}[t!]
\centering
\includegraphics[width=\linewidth]{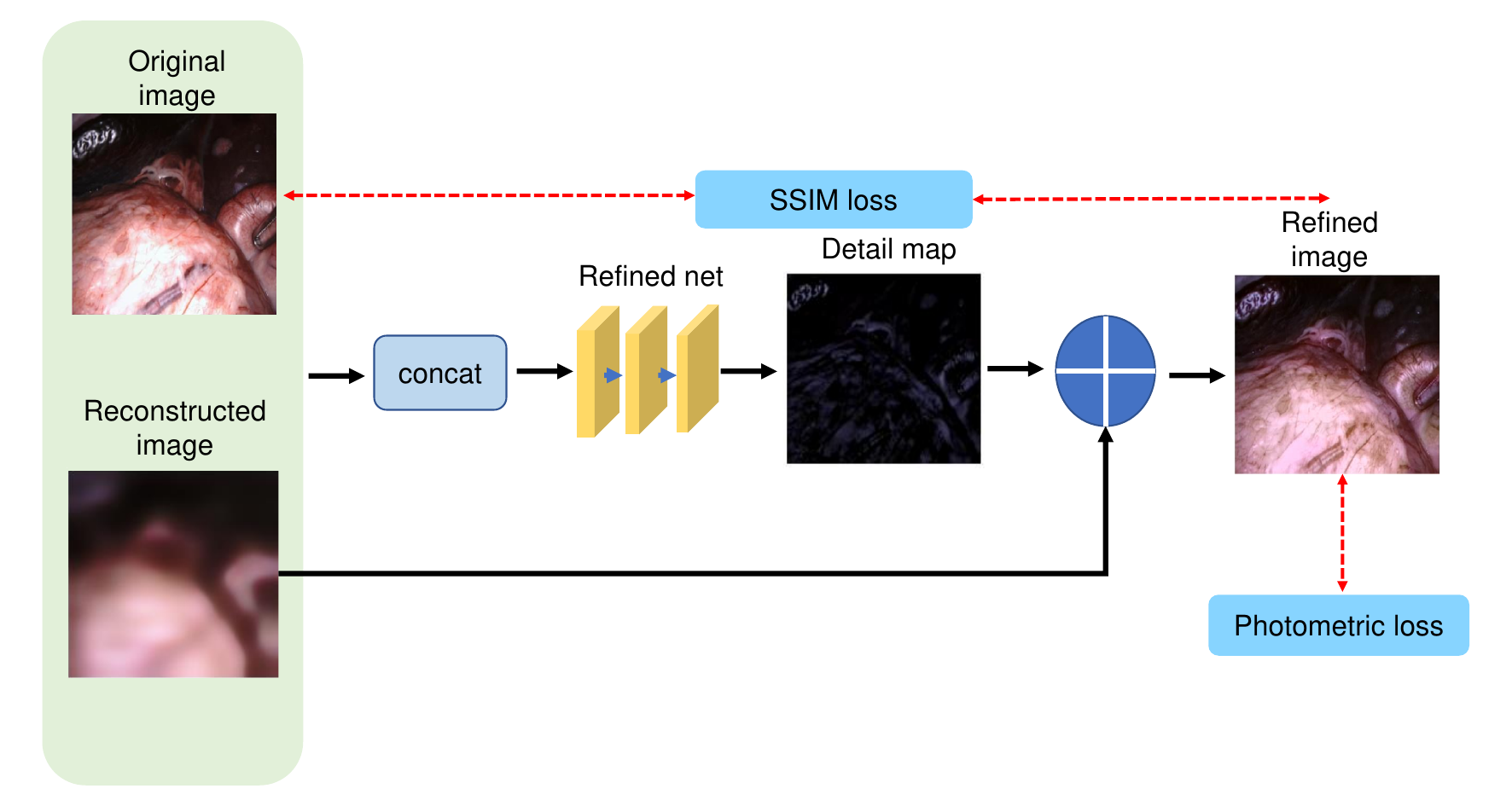}
\caption{The pipeline of refined net. Refined autonomously make a trade-off between original image and the reconsrtucted image. }
\vspace{-.5em}
\label{fig:refined} 
\end{figure}

As shown in \cref{fig:refined}, we design a neural network that autonomously selects between the features of the original and reconstructed images to generate optimal detail information. 
Specifically, the original image and reconstructed image are concated and fed to the refined net. Then, refined net autonomously strikes a balance and generate the optimal detail infomation. 
Finally, the optimal detail and the low-freauency reconstructed image are added to generate the refined image. 

In order to constrain the refined net to generate compatible features for depth and pose estimation, we use downstream photometric loss to optimize the refined net. In addition, we use the SSIM loss function to constrain the generated image and the original image so that the distribution of the generated image cannot be too far from that of the original image.

\subsubsection{refined network architecture}
Specificlly, our refined network employs en encoder-decoder structure with skip connections. In the encoding stage, we concat original image and reconstruction image and feed them through layers of convolutional blocks with a stride of 2, resulting in a five-level feature pyramid. Then, skip connections propagate the pyramid features into the decoding stage and produce the detail map. Finally, we apply the residual framwork to fuse reconstruction image and the detail map, to obtain the refined image.

\section{Experiments} 

\subsection{Datasets and Evaluation Metric} 

We evaluate our proposed module on three public datasets and one inhouse dataset:

\paragraph{\textbf{EndoSLAM Unity Dataset} } The EndoSLAM Unity dataset is acquired from synthetic software Unity with depth and pose ground truth. It contains three organs: colon, small intestine, and stomach. To evaluate the genrealization of mask module, we train network using small training data with supervision pipeline. Specifically, we split each organ datasets into three parts: first 1k frames for training, middle 1k frames for validation, and last 1k frames for testing.

\paragraph{\textbf{EndoSLAM Lowcam Dataset}} 
The EndoSLAM dataset is acquired from ex vivo porcine gastrointestinal tract organs and contains challenging ex vivo parts with only pose ground truths. It contains five organs: colon, small intestine, stomachI, stomachII, stomachIII, and each organs has five trajectories. Our experiment apply five-fold cross validation on it. 

\paragraph{\textbf{Scared Dataset}} 
The SCARED dataset is collected from fresh porcine cadaver abdominal anatomy and contains 35 endoscopic videos with the point cloud and ego-motion ground truth. We choose dataset1-4 as training set, and choose 1500 frames from dataset5-9 as test set for depth evaluation, and 18 trajectories for pose evaluation.

\subsection{Baselines} 

Since our proposed module is a plug-and-play module, we apply it to three classical baslines in depth and pose estimation networks to verify its effecicy. 
\begin{itemize}
    \item Monodepth v2:
    \item AF-SfMLearner: AF-SfMLearner is the official PyTorch implementation for training and testing depth estimation models using the method described in Self-Supervised Monocular Depth and Ego-Motion Estimation in Endoscopy: Appearance Flow to the Rescue1. The method is an extension of SfMLearner and uses appearance flow to handle non-rigid deformations.
    \item Endo-SfMLearner: Endo-SfMLearner is a self-supervised spatial attention-based monocular depth and pose estimation method. It is an unsupervised learning method that does not require ground truth depth data. The method combines residual networks with spatial attention module in order to dictate the network to focus on distinguishable and highly textured tissue regions. The brightness-aware photometric loss makes the predicted depth to be consistent under various illumination conditions
\end{itemize}

\subsection{Evaluation Metrics}
For depth estimation, following previous works, we adopt some standard evaluation metrics: absolute relative error(Abs Rel), the squared relative error (Sq Rel), root-mean-squared error (RMSE), root-mean-square logarithmic error(RMSE log), and the accuracy.

Following \cite{}, we use the median scaling to scale the predicted depth maps during evaluation. 
We denote the depth ground truth $\hat{D}$, and predicted depth map $\tilde{D}$, then the scaled depth map can be expressed as
\begin{equation}
\tilde{D}_{scaled} = \tilde{D} * median(\hat{D}) / median(\tilde{D})    
\end{equation}

\begin{table}[]
\centering
\begin{tabular}{c|c}
\hline
Metric   & Definition \\ \hline
Abs Rel  &  $\frac{1}{n}\sum_{i=1}^{n}|\hat{d}_{i} - d_{i}| / d_{i}$          \\ \hline
Sq Rel   &  $\frac{1}{n}\sum_{i=1}^{n}|\hat{d}_{i} - d_{i}|^{2} / d_{i}$           \\ \hline
RMSE     &  $\sqrt{\frac{1}{n}\sum_{i=1}^{n}|\hat{d}_{i} - d_{i}|^{2}}$          \\ \hline
RMS log  &   $\sqrt{\frac{1}{n}\sum_{i=1}^{n}|\log \hat{d}_{i} - \log d_{i}|^{2}}$          \\ \hline
Accuracy &    $N(\{ \frac{\hat{d}_{i}}{d_{i}}, \frac{d_{i}}{\hat{d}_{i}}\} < \delta) / n $       \\ \hline
\end{tabular}
\end{table}

For pose estimation, we adopt the 5-frame pose evaluation and the metric of absolute trajectory error(ATE) following previous works\cite{}.

\subsection{Depth performance}

\paragraph{Results on Scared Dataset}

\begin{table*}[tb]
\centering
\caption{ Depth performance comparison on SCARED dataset. We note that after incorporating our module, the performance of all three baselines improve by a large margin.
}
\label{table:scared_depth reuslts}
\resizebox{0.8\linewidth}{!}{
\begin{tabular}{cccccccc}
\toprule
 \multicolumn{1}{c|}{Method} & \multicolumn{1}{c|}{abs\_rel} & \multicolumn{1}{c|}{sq\_rel} & \multicolumn{1}{c|}{rmse} & \multicolumn{1}{c|}{rmse\_log} & \multicolumn{1}{c}{$a_{1}$}\\ 
\midrule
\midrule
 Monodepth v2               & 0.120$\pm$0.059           & 1.468$\pm$1.141                         & 8.297$\pm$3.547                         &0.147$\pm$0.064         & 0.859$\pm$0.129                  \\ \hline


\rowcolor{mygray}
 +mask\&refined                      & \textbf{0.101}$\pm$\textbf{0.049}                          & \textbf{1.098}$\pm$\textbf{1.019}                         & \textbf{7.058}$\pm$\textbf{3.254}                         & \textbf{0.129}$\pm$\textbf{0.058}                      & \textbf{0.905}$\pm$\textbf{0.108}                   \\ \hline
 AF-SfMLearner                & 0.105$\pm$0.051           & 1.183$\pm$1.083                         & 7.381$\pm$3.491                         &0.141$\pm$0.061         & 0.877$\pm$0.118                     \\ \hline

\rowcolor{mygray}
 +mask\&refined                       & \textbf{0.095}$\pm$\textbf{0.046}                          & \textbf{1.044}$\pm$\textbf{1.100}                        & \textbf{6.945}$\pm$\textbf{3.644}                         & \textbf{0.129}$\pm$\textbf{0.060}                      & \textbf{0.910}$\pm$\textbf{0.110}          \\ \hline
  
Endo-SfMLearner                & 0.119$\pm$0.056           & 1.462$\pm$1.096                         & 8.340$\pm$3.454                         &0.148$\pm$0.063         & 0.860$\pm$0.123                             \\ \hline


\rowcolor{mygray}
 +mask\&refined                       & \textbf{0.110}$\pm$\textbf{0.057}                          & \textbf{1.277}$\pm$\textbf{1.200}                        & \textbf{7.497}$\pm$\textbf{3.405}                         & \textbf{0.139}$\pm$\textbf{0.065}                      & \textbf{0.891}$\pm$\textbf{0.121}         \\ \hline
 
\bottomrule
\end{tabular}
}
\end{table*}

\cref{table:scared_depth reuslts} shows the single-view depth estimation results on Scared. Among three baselines, AF-SfMLearner achieves the best result, since it propose two additional network modules to alleviate the brightness inconsistency problem in endoscopic images.

One can see that after incorporating our proposed module, the performance of depth estimation is improved by a large margin.
It is noteworthy that the depth estimation performance of `\textbf{monodepth v2 + mask\&refined}' surpassed that of AF-SfMLearner, which demonstrates that images processed by our proposed module can unleash the power of the network better, even surpassing more advanced networks.

\begin{figure*}[]
\centering
\includegraphics[width=\linewidth]{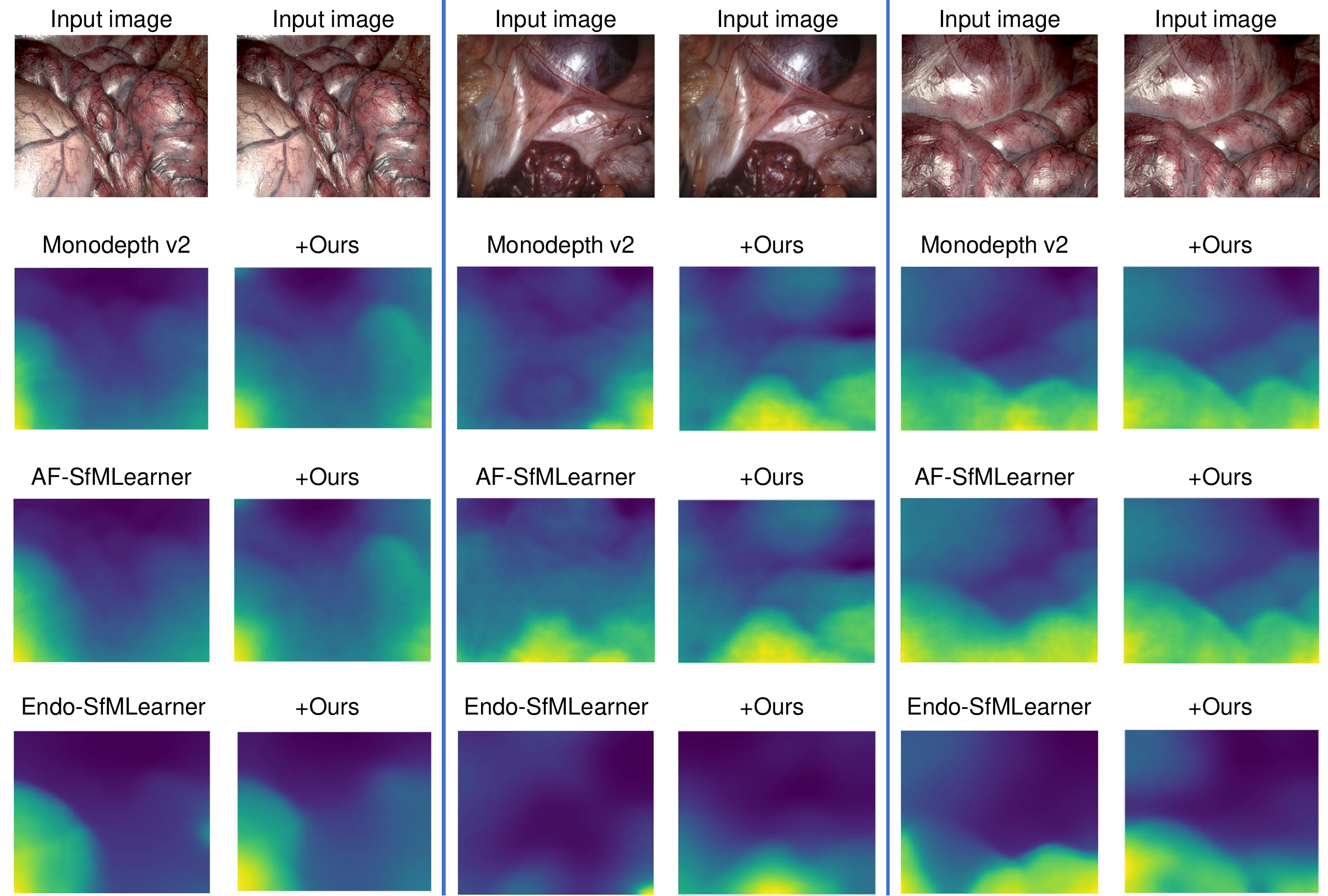}
\caption{ Qualitative depth comparison on the SCARED dataset. After incorporating our proposed module, the depth map is smoother and the structure is clearer.}
\vspace{-.5em}
\label{fig:scared_depth} 
\end{figure*}

See qualitative results of depth estimation in \cref{fig:scared_depth}. It is clear that our proposed method enables the network to estimate more stable depth information. For instance, for some dark-colored tissues in the second row, Monodepth v2 and Endo-SfMLEarner may identify them as distant background, but our method can accurately estimate the depth of the dark tissues.

\paragraph{Results on Lowcam Dataset} 

\begin{figure*}[]
\centering
\includegraphics[width=\linewidth]{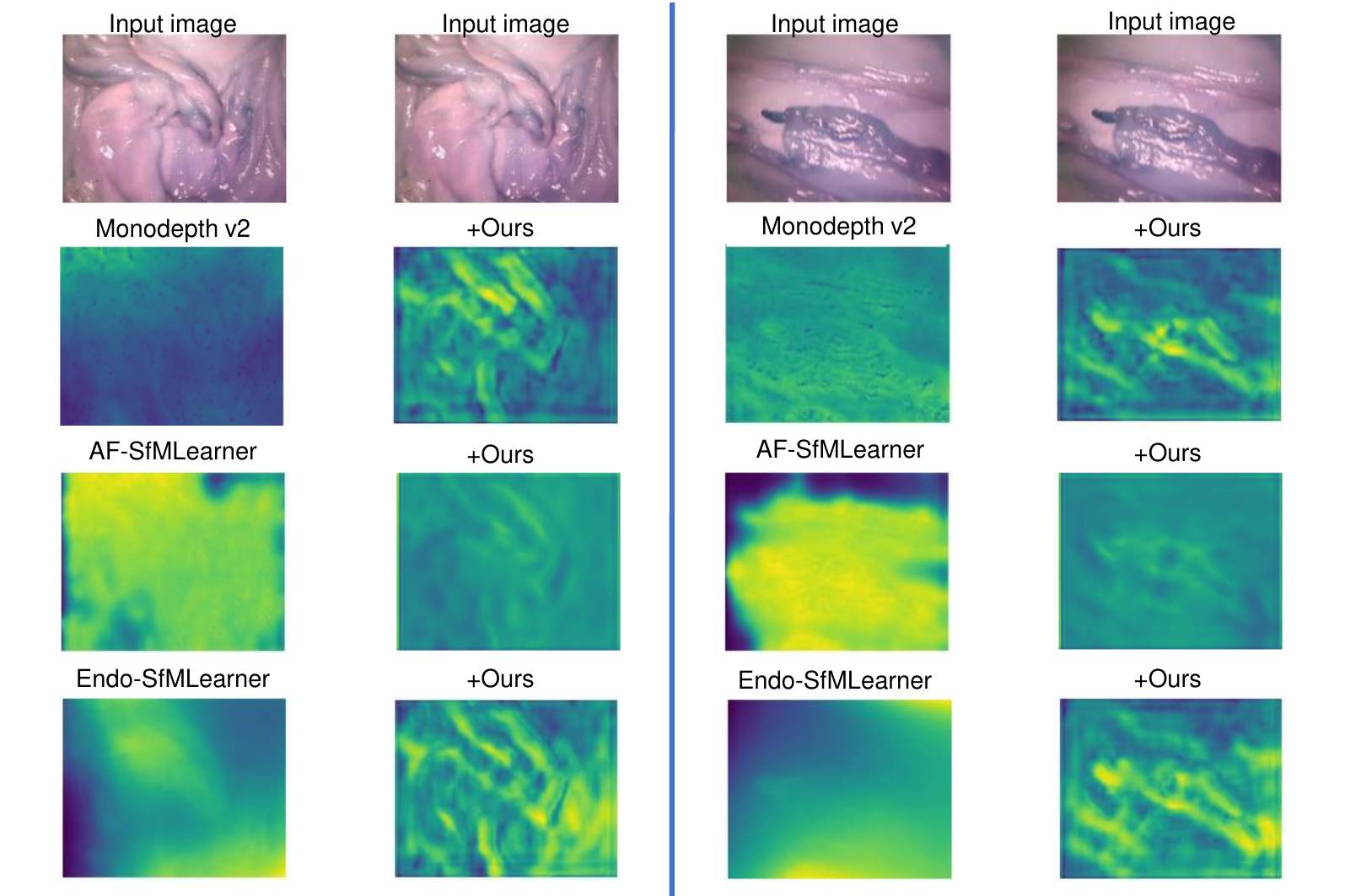}
\caption{ Qualitative depth comparison on the LOWCAM dataset. After incorporating our proposed module, the depth map is smoother and the structure is clearer.}
\vspace{-.5em}
\label{fig:lowcam_depth} 
\end{figure*}

Since lowcam dataset has no depth ground truth, we use qualitive evaluation. 
As shown in \cref{fig:lowcam_depth}, existing baselines cannot accurately estimate the depth, due to the artifact and sparse features. 
In contrast, after incorporating the module we proposed to the baseline, the network has a more global perception ability and the images have more stable features. Therefore, the network can estimate the overall trend of depth maps.

\paragraph{Results on inhouse Dataset}

\subsection{Pose performance}

\paragraph{Results on Scared Dataset} 

\cref{table:scared_pose result} shows the quantitative comparison on the SCARED dataset. It is clear that our proposed module shows significant improvement on three baselines. 

\begin{figure*}[]
\centering
\includegraphics[width=\linewidth]{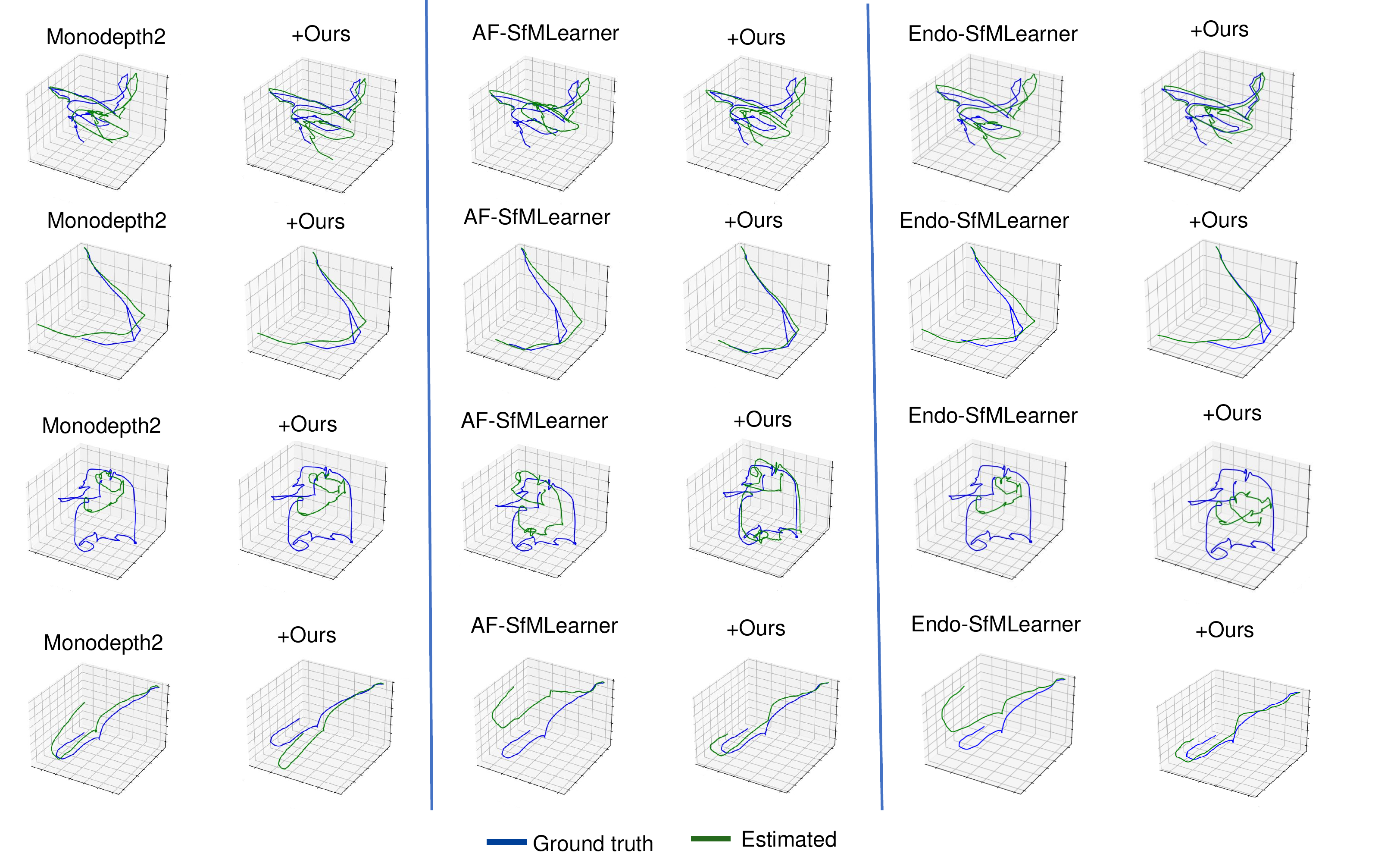}
\caption{ Qualitative pose comparison on the SCARED dataset. After incorporating our proposed module, the estimated trajectory is closer to ground truth.}
\vspace{-.5em}
\label{fig:scared_pose} 
\end{figure*}

The visual results of pose estimation are shown in the \cref{fig:scared_pose}.
The module we propose can make the predicted trajectory closer to the ground truth.
In addition, our module brings better scale consistency. In the third case in \cref{fig:scared_pose}, we can see that compared with AF-SfMLearner, the scale of ours is more consistent. 
This may be due to the large changes in the scene along the trajectory, resulting in the scale of AF inconsistent. Our enhanced images have more stable features and the train network perceives global information, resulting in a more stable perceptual ability, thus achieving a more consistent scale.

\begin{table}[!htb]
\centering
\caption{ Pose performance comparison on SCARED dataset. We
note that after incorporating our module, the performance of all
three baselines improve by a large margin}
\label{table:scared_pose result}
\centering
\begin{tabular}{cc}
\toprule
 \multicolumn{1}{c|}{Method} &\multicolumn{1}{c}{ATE} \\ 
\midrule
\midrule
 Monodepth v2             & 0.0805$\pm$0.0451          \\ \hline
 
\rowcolor{mygray}
 +mask\&refined               & \textbf{0.0678$\pm$0.0414}       \\ \hline

 AF-SfMlearner            & 0.0731$\pm$0.0451    \\ \hline

\rowcolor{mygray}
+mask\&refined               & \textbf{0.0648$\pm$0.0399}    \\ \hline

 Endo-SfMlearner            & 0.0731$\pm$0.0374    \\ \hline
\rowcolor{mygray}
 +mask\&refined            &\textbf{ 0.0687$\pm$0.0390 }   \\ \hline

\bottomrule
\end{tabular}

\end{table}

\paragraph{Results on Lowcam Dataset} 

EndoSLAM Lowcam dataset is a challenging dataset due to its sparse texture and challenging specular artifacts. 
We use 7 DoF align to align the predictive trajectory with the ground truth, to test whether the neurual network can predict the trajectory. 

\begin{figure*}[!htb]
\centering
\includegraphics[width=\linewidth]{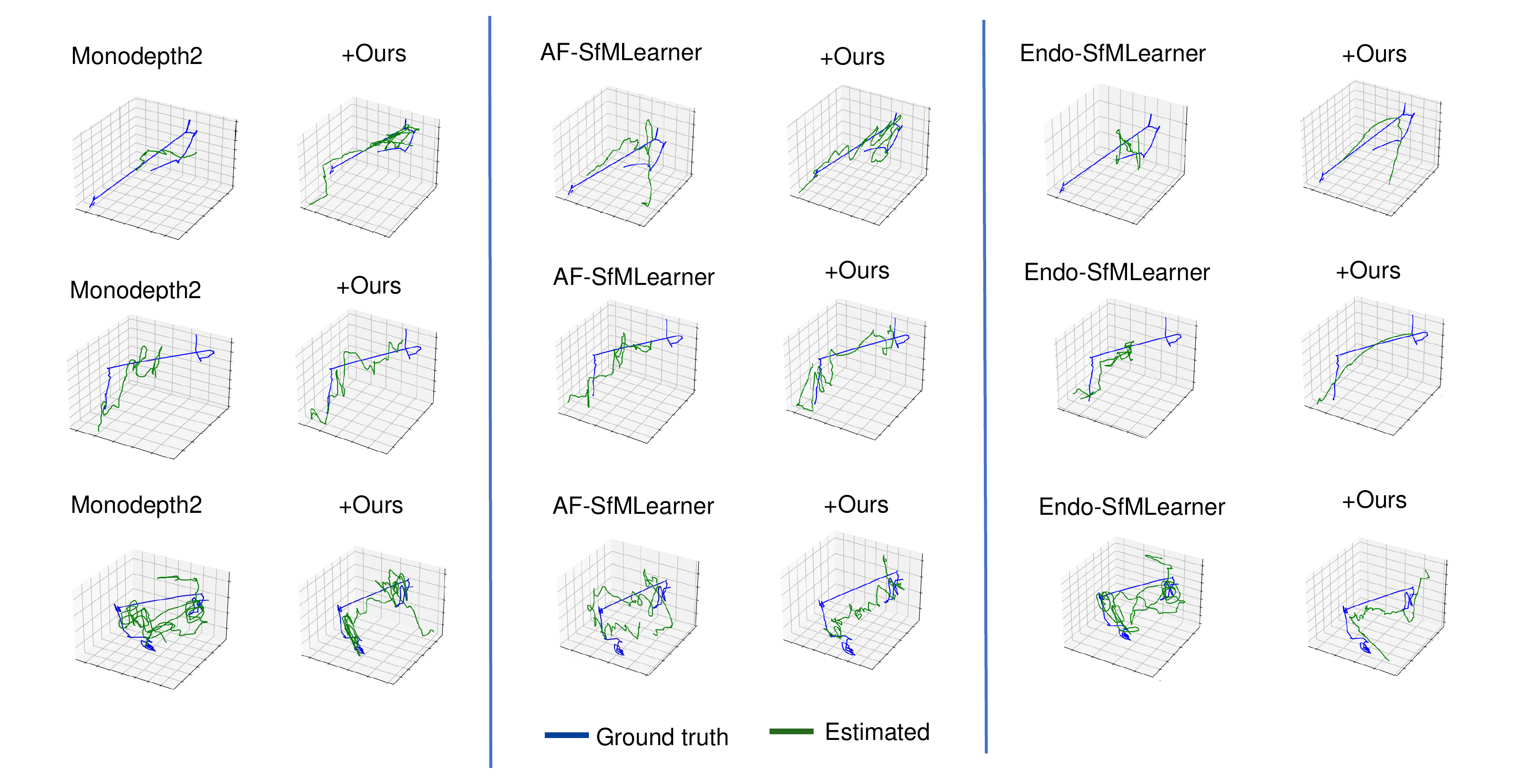}
\caption{ Qualitative pose comparison on the Lowcam dataset. After incorporating our proposed module, the estimated trajectory is closer to ground truth.}
\vspace{-.5em}
\label{fig:lowcam_pose} 
\end{figure*}

As shown in \cref{tab:lowcam pose}, compared with all baselines, our proposed module can achieve smaller ATE. And the visualization is shown in \cref{fig:lowcam_pose}. 
It is clear that our proposed module can provide better trajectory prediction.
We attribute this to our proposed enhance network, which optimizes images and is able to extract more stable features among various artifacts.

\begin{table*}[]
\caption{Pose performance comparison on LOWCAM dataset. We
note that after incorporating our module, the performance of all three baselines improve by a large margin}
\label{tab:lowcam pose}
\resizebox{\linewidth}{!}{
\begin{tabular}{cccccccc}
\hline
                                 & Method       & Traj1             & Traj2             & Traj3             & Traj4             & Traj5              & Avg. \\ \cline{3-8} 
                                 &              & ATE               & ATE               & ATE               & ATE               & ATE                & ATE \\ \hline
\multirow{6}{*}{Colon}           & Monodepth v2 & 0.1751$\pm$0.0741 & 0.0993$\pm$0.0529 & 0.0890$\pm$0.0477 & 0.0741$\pm$0.0348 & 0.0616$\pm$0.0399  & 0.0998$\pm$0.0499 \\
                                 & +Ours        & 0.1265$\pm$0.0833 & 0.0627$\pm$0.0616 & 0.0728$\pm$0.0355 & 0.0634$\pm$0.0306 & 0.0461$\pm$0.0311  & 0.0743$\pm$0.0484 \\ \cline{2-8} 
                                 & AF           & 0.1426$\pm$0.0683 & 0.1049$\pm$0.0616 & 0.7934$\pm$0.0473 & 0.0660$\pm$0.0321 & 0.05617$\pm$0.0404 & 0.0898$\pm$0.0499\\ 
                                 & +Ours        & 0.1450$\pm$0.0683 & 0.0642$\pm$0.0615 & 0.0776$\pm$0.0461 & 0.0505$\pm$0.0250 & 0.0516$\pm$0.0292  & 0.0777$\pm$0.0460\\ \cline{2-8} 
                                 & Endo-SfMLearner      & 0.2064$\pm$0.0885 & 0.1198$\pm$0.0755 & 0.1001$\pm$0.0625 & 0.0796$\pm$0.0428 & 0.0556$\pm$0.0363  & 0.1123$\pm$0.0611 \\ 
                                 & +Ours        & 0.1592$\pm$0.0791 & 0.0637$\pm$0.0557 & 0.0693$\pm$0.0374 & 0.0588$\pm$0.0343 & 0.0482$\pm$0.0290  & 0.0798$\pm$0.0471 \\ \hline
\multirow{6}{*}{Small Intestine} & Monodepth v2 & 0.0896$\pm$0.0447 & 0.0503$\pm$0.0589 & 0.0751$\pm$0.0277 & 0.0622$\pm$0.0262 & 0.1385$\pm$0.0765  & 0.0815$\pm$0.0468 \\ 
                                 & +Ours        & 0.0698$\pm$0.0466 & 0.0520$\pm$0.0591 & 0.0546$\pm$0.0245 & 0.0467$\pm$0.0284 & 0.1351$\pm$0.0787  & 0.0716$\pm$0.0476 \\ \cline{2-8} 
                                 & AF           & 0.0876$\pm$0.0493 & 0.0544$\pm$0.0530 & 0.0718$\pm$0.0243 & 0.0664$\pm$0.0212 & 0.1387$\pm$0.0779  & 0.0838$\pm$0.0451 \\
                                 & +Ours        & 0.0840$\pm$0.0495 & 0.0521$\pm$0.0591 & 0.0535$\pm$0.0329 & 0.0507$\pm$0.0252 & 0.1371$\pm$0.0796  & 0.0755$\pm$0.0493\\ \cline{2-8} 
                                 & Endo-SfMLearner      & 0.0922$\pm$0.0397 & 0.0519$\pm$0.0575 & 0.0564$\pm$0.0227 & 0.0666$\pm$0.0210 & 0.1363$\pm$ 0.0811 & 0.0812$\pm$0.0438  \\ 
                                 & +Ours        & 0.0792$\pm$0.0548 & 0.0540$\pm$0.0553 & 0.0496$\pm$0.0284 & 0.0413$\pm$0.0235 & 0.1368$\pm$0.0809  & 0.0722$\pm$0.0485 \\ \hline
\multirow{6}{*}{StomachI}        & Monodepth v2 & 0.0500$\pm$0.0389 & 0.1014$\pm$0.0928 & 0.0388$\pm$0.0285 & 0.0224$\pm$0.0103 &   ---              & 0.0532$\pm$0.0426 \\ 
                                 & +Ours        & 0.0505$\pm$0.0359 & 0.0924$\pm$0.0751 & 0.0364$\pm$0.0252 & 0.0187$\pm$0.0132 &   ---              & 0.0495$\pm$0.0373\\ \cline{2-8} 
                                 & AF           & 0.0512$\pm$0.0954 & 0.0954$\pm$0.0896 & 0.0387$\pm$0.0261 & 0.0243$\pm$0.0108 &   ---              & 0.0524$\pm$0.0554 \\ 
                                 & +Ours        & 0.0507$\pm$0.0342 & 0.0943$\pm$0.0929 & 0.0382$\pm$0.0252 & 0.0225$\pm$0.0184 &   ---              & 0.0520$\pm$0.0427 \\ \cline{2-8} 
                                 & Endo-SfMLearner      & 0.0502$\pm$0.0388 & 0.1012$\pm$0.0819 & 0.0396$\pm$0.0263 & 0.0279$\pm$0.0114 &   ---              & 0.0547$\pm$0.0396 \\
                                 & +Ours        & 0.0551$\pm$0.0387 & 0.1031$\pm$0.0955 & 0.0386$\pm$0.0247 & 0.0188$\pm$0.0126 &   ---              & 0.0539$\pm$0.0428 \\ \hline
\multirow{6}{*}{StomachII}       & Monodepth v2 & 0.0769$\pm$0.0521 & 0.0147$\pm$0.0014 & 0.0763$\pm$0.0484 & 0.0865$\pm$0.0411 &   ---              & 0.0636$\pm$0.0358 \\ 
                                 & +Ours        & 0.0718$\pm$0.0471 & 0.0129$\pm$0.0012 & 0.0727$\pm$0.0495 & 0.0869$\pm$0.0405 &   ---              & 0.0592$\pm$0.0519 \\ \cline{2-8} 
                                 & AF           & 0.0759$\pm$0.0601 & 0.0016$\pm$0.0017 & 0.0778$\pm$0.0452 & 0.0839$\pm$0.0368 &   ---              & 0.0598$\pm$0.0360 \\ 
                                 & +Ours        & 0.0750$\pm$0.0488 & 0.0013$\pm$0.0012 & 0.0616$\pm$0.0434 & 0.0770$\pm$0.0427 &   ---              & 0.0537$\pm$0.0417 \\ \cline{2-8} 
                                 & Endo-SfMLearner      & 0.0823$\pm$0.0572 & 0.0016$\pm$0.0017 & 0.0781$\pm$0.0470 & 0.0808$\pm$0.0353 &   ---              & 0.0607$\pm$0.0353 \\ 
                                 & +Ours        & 0.0689$\pm$0.0421 & 0.0013$\pm$0.0013 & 0.0740$\pm$0.0456 & 0.0787$\pm$0.0452 &   ---              & 0.0557$\pm$0.0355 \\ \hline
\multirow{6}{*}{StomachIII}      & Monodepth v2 & 0.0970$\pm$0.0770 & 0.1101$\pm$0.0509 & 0.1035$\pm$0.0840 & 0.2581$\pm$0.1140 &   ---              & 0.1422$\pm$0.0815 \\ \cline{8-8} 
                                 & +Ours        & 0.0968$\pm$0.0788 & 0.1070$\pm$0.0510 & 0.1036$\pm$0.0857 & 0.2513$\pm$0.1066 &   ---              & 0.1396$\pm$0.0802 \\ \cline{2-8} 
                                 & AF           & 0.0993$\pm$0.0780 & 0.1061$\pm$0.0461 & 0.1193$\pm$0.0631 & 0.2236$\pm$0.1064 &   ---              & 0.1371$\pm$0.0734 \\  
                                 & +Ours        & 0.0945$\pm$0.0801 & 0.1030$\pm$0.0532 & 0.0968$\pm$0.0837 & 0.2230$\pm$0.1021 &   ---              & 0.1293$\pm$0.0797 \\ \cline{2-8} 
                                 & Endo-SfMLearner      & 0.1001$\pm$0.0722 &      0.0996±0.0499             & 0.1184$\pm$0.0570 & 0.2430$\pm$0.0999 &   ---              & 0.1419$\pm$0.0688 \\  
                                 & +Ours        & 0.0975$\pm$0.0695 & 0.1083$\pm$0.0459 & 0.0959$\pm$0.0807 & 0.2491$\pm$0.1115 &   ---              & 0.1377$\pm$0.0769 \\ \hline
\end{tabular}
}
\end{table*}

\section{Ablation study} 

\subsection{Explored paradigm} 

The effectiveness of our method depends on two components, i.e. a). MIM module with reconstruction decoder; b). refined module. To invenstigate the influence of each component on the depth and pose estimation, we conduct an ablation study. 
The result is shown in \cref{table:ablation}. 
We can see that monodepth v2 with none of our contributions, performs poorly.
The mask module and refined net can improve the performance, respectively.
With all of the contributions combined, we obtain the best results.

\begin{table*}[]
\caption{\textbf{Ablation on each component.} We use Monodepth v2 as baseline model and evaluate on SCARED dataset. We find that each component plays a critical role.}
\label{table:ablation}
\centering
\resizebox{\linewidth}{!}{
\begin{tabular}{ccc|ccccc}
\hline
\multicolumn{3}{c|}{Component} & \multirow{2}{*}{abs\_rel} & \multirow{2}{*}{sq\_rel} & \multirow{2}{*}{rmse} & \multirow{2}{*}{rmse\_log} & \multirow{2}{*}{$a_{1}$}\\ \cline{1-3}
MIM  & Reconstruction decoder   &    Refined net &                          \\ \hline
 \textcolor{light-gray}{\XSolidBrush}          &     \textcolor{light-gray}{\XSolidBrush}   &   \textcolor{light-gray}{\XSolidBrush}    & 0.120$\pm$0.059 & 1.468$\pm$1.141 & 8.297$\pm$3.547 & 0.147$\pm$0.064 & 0.859$\pm$0.129                         \\ \hline
\CheckmarkBold        &   \textcolor{light-gray}{\XSolidBrush}     &   \textcolor{light-gray}{\XSolidBrush}   &  0.116$\pm$0.067 & 1.486$\pm$1.613 & 7.964$\pm$4.005 & 0.145$\pm$0.073 & 0.873$\pm$0.137                       \\ \hline

\CheckmarkBold          &      \CheckmarkBold     &  \textcolor{light-gray}{\XSolidBrush}  & 0.116$\pm$0.054 & 1.274$\pm$1.125 & 7.386$\pm$3.419 & 0.136$\pm$0.063 & 0.892$\pm$0.113                         \\ \hline
\CheckmarkBold         &       \CheckmarkBold     &  \CheckmarkBold  & 0.101$\pm$0.049 & 1.098$\pm$1.019 & 7.058$\pm$3.254 & 0.129$\pm$0.058 & 0.905$\pm$0.108                         \\ \bottomrule

\end{tabular}}
\end{table*}

\subsection{Different mask strategies} 

Adopting different mask strategies will have varying effects on the results. Therefore, in this section, we conducted detailed experiments on mask strategies to find the optimal mask strategy.

Specifically, we conduct two ablation studies, first is the choice of mask strategies, including random mask, artifact-suppressing mask, and uncertainty mask. 
The second is the ratio of mask area.

The results are shown in \cref{tab:mask strategy}. Firstly, we found that all mask strategies can improve the performance of depth estimation. Secondly, the mask based on artifact suppressing achieves the best result, while the random mask performed the worst. In addition, we found that the mask based on uncertainty and the mask based on brightness are complementary. Combining the two can achieve better results.

\begin{table*}[]
\label{tab:mask strategy}
\caption{Ablation study on the optimal strategy of masking. }
\resizebox{\linewidth}{!}{
\begin{tabular}{cccccccc}
\hline
                            mask choice   & mask ratio       & abs\_rel             & sq\_rel            & rmse             & rmse\_log             & $a_{1}$              &  \\ \hline
\multirow{3}{*}{Random mask}           &20\% & 0.1751$\pm$0.0741 & 0.0993$\pm$0.0529 & 0.0890$\pm$0.0477 & 0.0741$\pm$0.0348 & 0.0616$\pm$0.0399   \\ 
                                 &40\%       & 0.1265$\pm$0.0833 & 0.0627$\pm$0.0616 & 0.0728$\pm$0.0355 & 0.0634$\pm$0.0306 & 0.0461$\pm$0.0311  
                                 \\ 
                                 &60\%        & 0.1450$\pm$0.0683 & 0.0642$\pm$0.0615 & 0.0776$\pm$0.0461 & 0.0505$\pm$0.0250 & 0.0516$\pm$0.0292   \\ \hline

\multirow{3}{*}{bright mask}           &20\% & 0.1751$\pm$0.0741 & 0.0993$\pm$0.0529 & 0.0890$\pm$0.0477 & 0.0741$\pm$0.0348 & 0.0616$\pm$0.0399   \\  
                                 &40\%       & 0.1265$\pm$0.0833 & 0.0627$\pm$0.0616 & 0.0728$\pm$0.0355 & 0.0634$\pm$0.0306 & 0.0461$\pm$0.0311 
                                 \\ 
                                 &60\%        & 0.1450$\pm$0.0683 & 0.0642$\pm$0.0615 & 0.0776$\pm$0.0461 & 0.0505$\pm$0.0250 & 0.0516$\pm$0.0292    \\ \hline

\multirow{3}{*}{uncertainty mask}           &20\% & 0.1751$\pm$0.0741 & 0.0993$\pm$0.0529 & 0.0890$\pm$0.0477 & 0.0741$\pm$0.0348 & 0.0616$\pm$0.0399 \\ 
                                 &40\%       & 0.1265$\pm$0.0833 & 0.0627$\pm$0.0616 & 0.0728$\pm$0.0355 & 0.0634$\pm$0.0306 & 0.0461$\pm$0.0311 
                                 \\ 
                                 &60\%        & 0.1450$\pm$0.0683 & 0.0642$\pm$0.0615 & 0.0776$\pm$0.0461 & 0.0505$\pm$0.0250 & 0.0516$\pm$0.0292  \\ \hline 

\multirow{3}{*}{bright + uncertainty mask}           &20\% & 0.1751$\pm$0.0741 & 0.0993$\pm$0.0529 & 0.0890$\pm$0.0477 & 0.0741$\pm$0.0348 & 0.0616$\pm$0.0399 \\ 
                                 &40\%       & 0.1265$\pm$0.0833 & 0.0627$\pm$0.0616 & 0.0728$\pm$0.0355 & 0.0634$\pm$0.0306 & 0.0461$\pm$0.0311 
                                 \\ 
                                 &60\%        & 0.1450$\pm$0.0683 & 0.0642$\pm$0.0615 & 0.0776$\pm$0.0461 & 0.0505$\pm$0.0250 & 0.0516$\pm$0.0292  \\ \hline

\end{tabular}
}
\end{table*}

In addition, we find that the mask can forbid the effect of many aratifacts.
\section{Discussion} 

In addition to improving the performance of unsupervised endoscopic depth pose estimation, the module we propose also performs well in other tasks.

\subsection{Generalization ability of mask module}

It is interesting that our proposed mask module can be a good data augmentation method. 
We evaluate it on supervised depth estimation task. 

We conducted tests on the EndoSLAM Unity dataset. This dataset is a computer-generated virtual dataset with accurate depth ground truth. For each sub-organ, we randomly selected 1000 images as the training set and used the remaining images as the test set. We used the Unet network architecture and the L2 loss function.

To validate the generalization ability of our mask module, we consider mask module as a data augmentation method and compare it with other data augmentation baselines, e.g., CDA and PDA. 

CDA is widely used data augmentation method like random-flipping and random-cropping. 

PDA is a data augmentation for depth estimation, which perturbs the camera pose and reprojects the image and corresponding depth ground truth.

First, we train and test the depth estimation network on the same dataset to validate the generalization ability on test set. 

The result is shown in \cref{tab:unity}.Compared to data augmentation using CDA and PDA, using MIM as a data augmentation method has stronger generalization in endoscopic scenarios.

\begin{table*}[!htb]
\caption{Depth performance trained and tested on the same organ }
\label{tab:unity}
\resizebox{\linewidth}{!}{
\begin{tabular}{cccccccc}
\hline
                            Organ   & Method       & abs\_rel             & sq\_rel            & rmse             & rmse\_log             & $a_{1}$              &  \\ \hline
\multirow{4}{*}{Colon}           &no aug & 0.130$\pm$0.026 & 0.219$\pm$0.182 & 1.469$\pm$0.853 & 0.174$\pm$0.036 & 0.839$\pm$0.089   \\ 
                                 &CDA & 0.123$\pm$0.018 & 0.177$\pm$0.081 & 1.299$\pm$0.541 & 0.165$\pm$0.027 & 0.865$\pm$0.058   \\ 
                                 &PDA       & 0.134$\pm$0.025 & 0.224$\pm$0.120 & 1.549$\pm$0.653 & 0.188$\pm$0.036 & 0.818$\pm$0.099  
                                 \\ 
                                 &Mask        & 0.117$\pm$0.022 & 0.162$\pm$0.080 & 1.214$\pm$0.030 & 0.162$\pm$0.030 & 0.877$\pm$0.067   \\ \hline

\multirow{4}{*}{Small intestine} &no aug & 0.146$\pm$0.097 & 1.528$\pm$2.794 & 6.002$\pm$4.056 & 0.186$\pm$0.038 & 0.853$\pm$0.132   \\           
                                 &CDA & 0.107$\pm$0.064 & 0.888$\pm$1.755 & 5.171$\pm$3.452 & 0.144$\pm$0.063 & 0.920$\pm$0.078   \\  
                                 &PDA       & 0.097$\pm$0.054 & 0.815$\pm$1.357 & 5.229$\pm$3.389 & 0.136$\pm$0.057 & 0.927$\pm$0.080 
                                 \\ 
                                 &Mask        & 0.095$\pm$0.054 & 0.782$\pm$1.357 & 4.678$\pm$3.379 & 0.133$\pm$0.061 & 0.936$\pm$0.076    \\ \hline

\multirow{4}{*}{Stomach}         &no aug & 0.248$\pm$0.062 & 2.141$\pm$1.666 & 7.021$\pm$4.218 & 0.316$\pm$0.109 & 0.541$\pm$0.155   \\ 
                                 &CDA & 0.240$\pm$0.077 & 1.949$\pm$1.271 & 5.806$\pm$2.648 & 0.268$\pm$0.078 & 0.674$\pm$0.159 \\ 
                                 &PDA       & 0.201$\pm$0.082 & 1.615$\pm$1.461 & 5.816$\pm$3.724 & 0.254$\pm$0.099 & 0.717$\pm$0.195 
                                 \\ 
                                 &Mask        & 0.178$\pm$0.064 & 1.233$\pm$1.205 & 4.402$\pm$2.292 & 0.224$\pm$0.071 & 0.821$\pm$0.120  \\ \hline 

\end{tabular}
}
\end{table*}

The visualization result is shown as XXX.

\subsection{Advantages of enhanced image} 

In this section, we will analyze the advantages of the enhanced images generated by our refined module, which autonomously selects the optimal details for image enhancement through a neural network.

Firstly, we find that image enhancement can significantly alleviate the illumination changes in endoscopic images. Illumination changes can lead to errors in theoretical priors and estimation errors in pose estimation based on the SfM framework. Therefore, when the enhancement module selects image details through the downstream photometric loss function, it will suppress some of the illumination details, thereby significantly alleviating the illumination changes in endoscopic images.

\begin{figure}[]
\centering
\includegraphics[width=\linewidth]{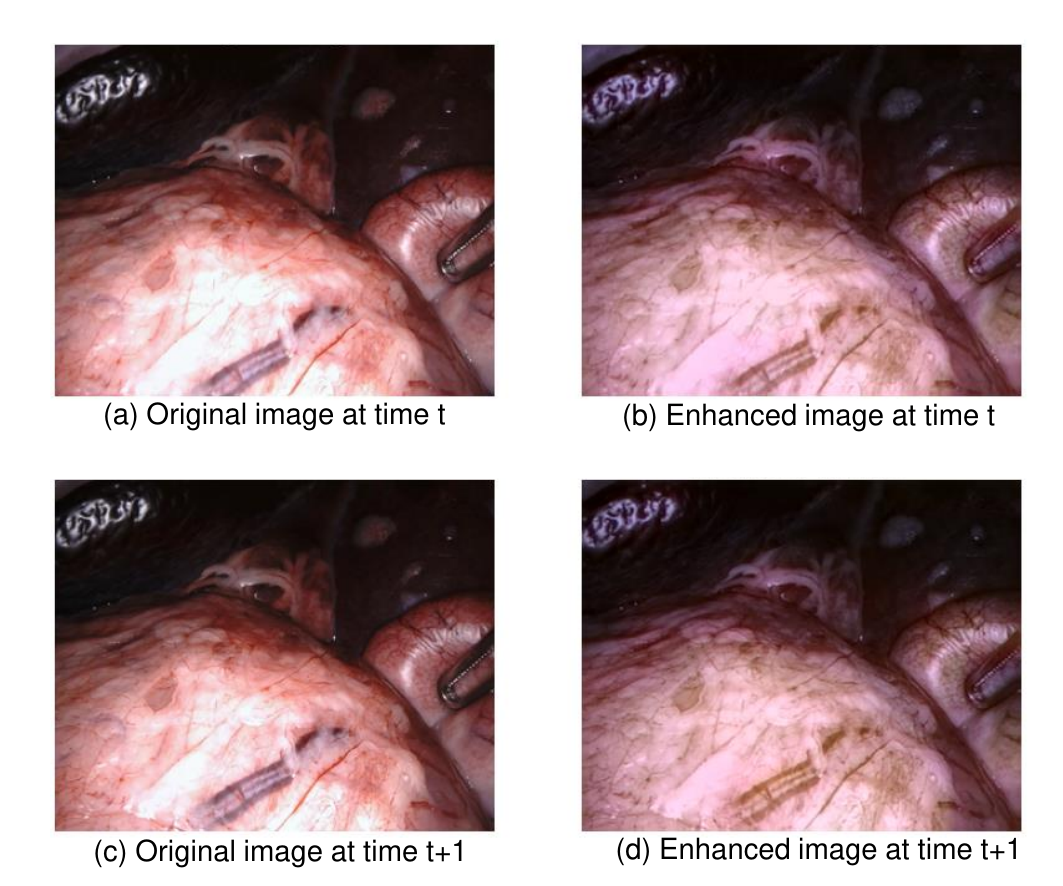}
\caption{ Comparison of two consecutive frames of the original and enhanced images. It is clear that the enhanced image has suppressed the characteristics of light variation.”}
\vspace{-.5em}
\label{fig:suppress_highlight} 
\end{figure}

Secondly, since the enhancement network is optimized through photometric loss, the images generated by the enhancement network have better adaptability with the depth and pose estimation network. Taking Monodepth v2 as an example, we maintained the consistency of the network structure and training process, and simply replaced the training images with enhanced images, which greatly improved the performance of depth and pose estimation, as shown in \cref{table:different image}. 

\begin{table*}[!htb]
\centering
\caption{ Depth performance comparison using different training images on SCARED dataset. We note that after simply replacing training images with refined images, the depth performance improves by a large margin.
}
\label{table:different image}
\resizebox{0.8\linewidth}{!}{
\begin{tabular}{cccccccc}
\toprule
 \multicolumn{1}{c|}{Training images} & \multicolumn{1}{c|}{abs\_rel} & \multicolumn{1}{c|}{sq\_rel} & \multicolumn{1}{c|}{rmse} & \multicolumn{1}{c|}{rmse\_log} & \multicolumn{1}{c}{$a_{1}$}\\ 
\midrule
\midrule
 Original images               & 0.120$\pm$0.059           & 1.468$\pm$1.141                         & 8.297$\pm$3.547                         &0.147$\pm$0.064         & 0.859$\pm$0.129                  \\ \hline


\rowcolor{mygray}
refined images                     & \textbf{0.113}$\pm$\textbf{0.066}                          & \textbf{1.426}$\pm$\textbf{1.077}                         & \textbf{8.104}$\pm$\textbf{3.429}                         & \textbf{0.143}$\pm$\textbf{0.060}                      & \textbf{0.874}$\pm$\textbf{0.146}                   \\ \hline

\bottomrule
\end{tabular}
}
\end{table*}

Finally, we find that refined images can extract more stable features. 
We conducted an experiment on feature point matching to prove that the generated augmented image features are more stable than the original image. 
First, we randomly selected 100 pairs of endoscopic images. Then, we used five feature point extraction algorithms, i.e., ORB, SURF, SIFT, SuperPoint, D2Net, to extract the feature points of the image, and then matched the features through the KNN algorithm.
Finally, we calculate the repeatability as an evaluation index for the correct matching rate.

\begin{table}[!htb]
\caption{Repeatability of five methods. For all feature extraction algorithms, repeatability is improved when using enhanced images.
}
\label{table:keypoint}
\resizebox{\linewidth}{!}{
\begin{tabular}{ccc}
\toprule
 \multicolumn{1}{c|}{Method} & \multicolumn{1}{c|}{Original image}  & \multicolumn{1}{c}{Refined image}\\ 
\midrule
\midrule
 ORB               & 0.525           & 0.677    \\ 
 SURF              & 0.590           & 0.663     \\ 
 SIFT              & 0.171           & 0.368     \\ 
 SuperPoint        & 0.628           & 0.693     \\ 
 D2Net              & 0.609          & 0.700    \\ 

\bottomrule
\end{tabular}
}
\end{table}

\cref{table:keypoint} shows the experimental result. It is clear that for all keypoint extraction algorithms, refined image can provide stabler features, then improve the repearability.
\section{Conclusion}

{\small
\bibliographystyle{unsrt}
\bibliography{egbib}
}

\end{document}